\pdfoutput = 1
\documentclass[10pt]{article}
\usepackage{spconf,amsmath,graphicx}
\usepackage[table,xcdraw]{xcolor}
\usepackage[super]{nth}

\title{Predicting Video features from EEG and Vice versa}
\name{Gautam Krishna \qquad Co Tran \qquad Mason Carnahan \qquad Ahmed H Tewfik}
\address{Brain Machine Interface Lab, The University
of Texas at Austin \\}
\begin{document}
%
\maketitle
\begin{abstract}
In this paper we explore predicting facial or lip video features from electroencephalography (EEG) features and predicting EEG features from recorded facial or lip video frames using deep learning models. The subjects were asked to read out loud English sentences shown to them on a computer screen and their simultaneous EEG signals and facial video frames were recorded.
Our model was able to generate very broad characteristics of the facial or lip video frame from input EEG features. Our results demonstrate the first step towards synthesizing high quality facial or lip video from recorded EEG features. We demonstrate results for a data set consisting of seven subjects. 
\end{abstract}
\begin{keywords}
Electroencephalography (EEG), deep learning, computer vision, video processing 
\end{keywords}
\section{Introduction}
\label{sec:intro}

The electroencephalography (EEG) is a non-invasive technique for measuring electrical activity of human brain where EEG sensors are placed on the scalp of the subject to obtain the EEG recordings. The EEG signals reflects electrical activity of millions of synchronous cortical neurons sharing similar spatial orientation. EEG offers very high temporal resolution even though the spatial resolution and signal to noise ratio offered are poor. The non-invasive nature of EEG makes it easy to study and deploy compared to other invasive neural recording techniques like electrocorticography (ECoG) and local field potentials. EEG signals are commonly used to drive various brain computer interface (BCI) applications. For example in references \cite{krishna20,krishna2019speech} authors show that EEG signals can be used to perform continuous and silent speech recognition where the EEG signals recorded in parallel with speech are translated to text. In \cite{krishna2020towards} authors demonstrate continuous silent speech recognition where they translated EEG signals recorded in parallel while subjects were silently reading English sentences in their mind to text.
Similarly in \cite{krishna2020advancing,krishna2020synthesis} authors provided preliminary results for synthesizing speech from EEG features. 

In \cite{krishna20,krishna2019speech} authors also demonstrated that EEG features can be used to enhance the performance of automatic speech recognition (ASR) systems operating in presence of background noise. The references \cite{assael2016lipnet,xu2018lcanet,shillingford2018large,afouras2018deep,petridis2018audio} demonstrated continuous audio-visual speech recognition and end-to-end lip reading. Technologies like speech recognition using EEG, lip reading can help people who can't produce voice or people with speaking disabilities to use virtual personal assistants like Amazon Alexa, Apple Siri etc there by improving technology accessibility.

The performance of visual speech recognition systems and lip reading systems degrades in presence of darkness and performance of audio-visual speech recognition systems degrades in presence of background noise. In \cite{krishna2019continuous} authors demonstrated that EEG features can be used to improve the performance of visual and audio-visual speech recognition systems. In this paper we study the problem of predicting facial video features from recorded EEG features and it's inverse problem, ie: predicting EEG features from recorded video frames.
We make use of the data set used by authors in \cite{krishna2019continuous} for this work and we demonstrate our results for seven subjects during test time. 

Our deep learning model was able to generate facial video frame from input EEG features with very broad characteristics and our results demonstrate the first step towards the end goal of synthesizing high quality video frames from EEG features. Generating facial frames from neural EEG signals and vice-versa might help in better understanding the underlying neuroscience principles behind lip reading, facial expressions etc.

\section{Deep Learning Models}
\label{sec:format}

The Figure 1 explains the architecture of the deep learning model used to predict video from input EEG features. The model takes EEG features of the shape [batch size, time steps, 30] as input and produces video of shape [batch size,time steps,100,100] as output. The temporal convolutional network (TCN) \cite{bai2018empirical} layer had 128 filters, the time distributed dense layers contained linear activation functions. The first time distributed dense layer contained 10000 hidden units and the final time distributed dense layer contained 100 hidden units.  The first time distributed dense layer's output is reshaped to shape [batch size,time steps, 100,100]. Each of the two dimensional convolutional transpose layers consists of 100 filters with a kernel size equal to (1,1) and rectified linear unit (ReLU) activation function. The two dimensional convolutional transpose layer outputs are passed to a two dimensional up-sampling layer with size equal to (1,1). The model was trained for 500 epochs using adam \cite{kingma2014adam} as the optimizer with the batch size set to 100. We used mean squared error (MSE) as the loss function and the validation split hyper parameter was set to a value of 0.05. Figure 3 shows the training and validation loss. 

The Figure 2 explains the architecture of the deep learning model used to predict EEG features from input video frames. 
The model takes video of shape [batch size,time steps,100,100] as input and produces 
EEG features of the shape [batch size, time steps, 30]  as output. Each of the two dimensional convolutional layers had 100 filters with kernel size equal to (1,3) and ReLU activation function. The two dimensional max pooling layer had a pool size of (1,2). After flattening the max pool layer output it is reshaped to the shape [batch size, time steps, shape of flatten layer[1]/time steps]. The time distributed dense layer consists of 30 hidden units and linear activation function. The model was trained for 1000 epochs with adam as the optimizer. The batch size was set to 100 and the validation split hyper parameter was set to 0.05. We used MSE as the loss function. 

For each subject for each experiment we used 10\% of the data as test set, 85\% of the data as training set and 5\% as validation set. The train-test-validation split was done randomly and there was no overlap between training, test and validation set. 

\begin{figure}[h]
\begin{center}
\includegraphics[height=6cm,width=0.3\textwidth,trim={0.1cm 0.1cm 0.1cm 0.1cm},clip]{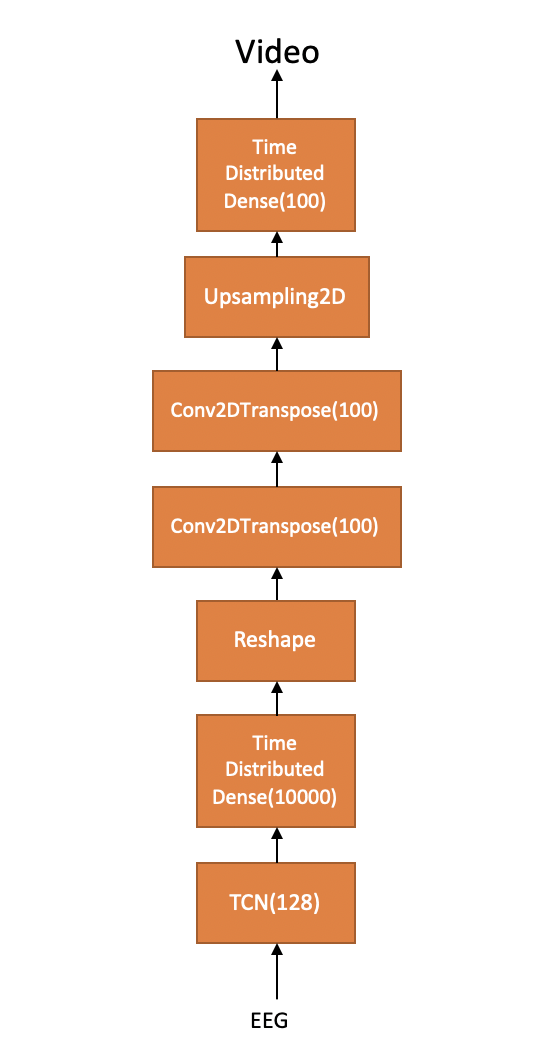}
\caption{Model for predicting video frames from EEG features} 
\label{1vsall}
\end{center}
\end{figure}

\begin{figure}[h]
\begin{center}
\includegraphics[height=6cm,width=0.3\textwidth,trim={0.1cm 0.1cm 0.1cm 0.1cm},clip]{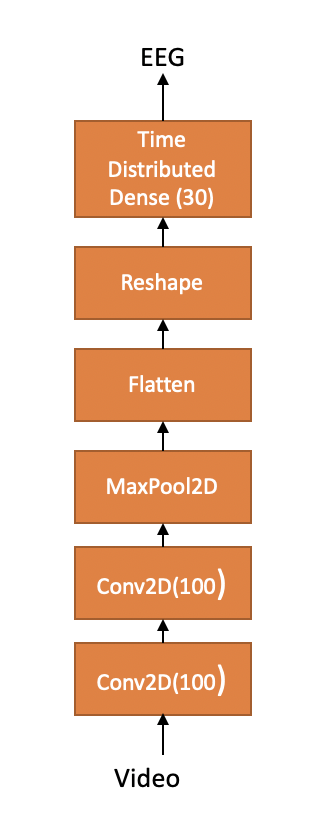}
\caption{Model for predicting EEG features from video frames} 
\label{1vsall}
\end{center}
\end{figure}

\begin{figure}[h]
\begin{center}
\includegraphics[height=5cm,width=0.3\textwidth,trim={0.1cm 0.1cm 0.1cm 0.1cm},clip]{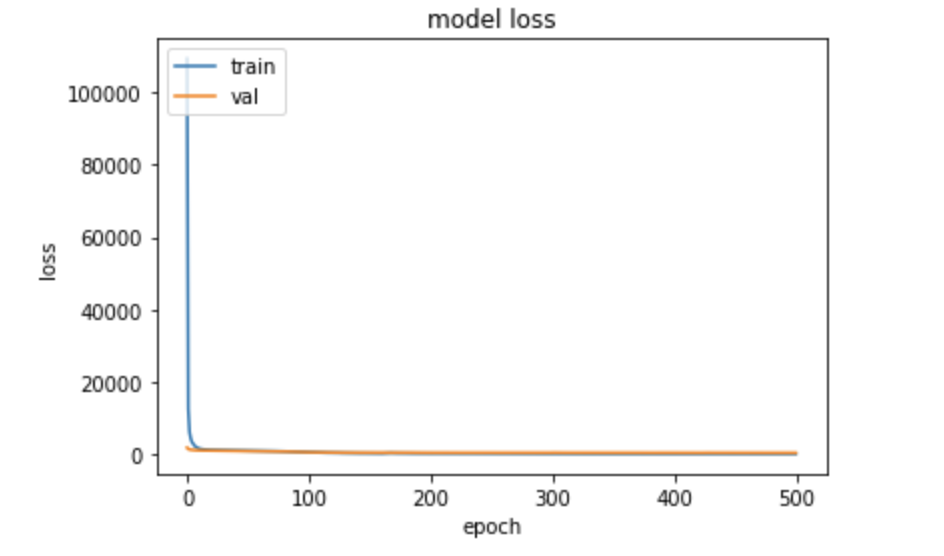}
\caption{Training and Validation loss for model used to predict video from EEG features} 
\label{1vsall}
\end{center}
\end{figure}

\section{Data Sets used for performing experiments}
\label{sec:typestyle}

We used the data set used by authors in \cite{krishna2019continuous} for this work. It consists of simultaneous recordings of EEG signals and facial video of subjects uttering English sentences. The data set consists of simultaneous EEG and video data from seven subjects. More details of experiment design, data set, EEG recording hardware are explained in \cite{krishna2019continuous}. 

\section{EEG and video feature extraction details}
\label{sec:majhead}

We followed the same EEG and video preprocessing techniques used by authors in \cite{krishna2019continuous} to process the EEG and video data. 

The EEG signals were sampled at 1000Hz and a fourth order IIR band pass filter with cut off frequencies 0.1Hz and 70Hz was applied. A notch filter with cut off frequency 60 Hz was used to remove the power line noise.
EEGlab's \cite{delorme2004eeglab} Independent component analysis (ICA) toolbox was used to remove other biological signal artifacts like electrocardiography (ECG), electromyography (EMG), electrooculography (EOG) etc from the EEG signals.

Then we extracted five statistical features for EEG, namely root mean square, zero crossing rate,moving window average,kurtosis and power spectral entropy \cite{krishna2019speech,krishna20}.  In total there were 155 features (31(channels) X 5) for EEG signals. The EEG features were extracted at a sampling frequency of 100Hz for each EEG channel.

Like explained by authors in \cite{krishna2019continuous} we extracted 100 frames per second from the recorded video. We used YOLO\cite{redmon2016you} object recognition model to perform face recognition from the extracted video frames. Then all RGB face frames were transformed to gray scale and then we resized all the gray scale face frames to a dimension of 100 X 100 using python imaging library (PIL). We further extracted lip or mouth frames from the gray scale face frames using DLib and iBug face landmark predictor with 68 landmarks \cite{sagonas2013300}. The iBug face landmark predictor was not able to detect mouth or lip for all the face frames, hence we used a mixture of facial and lip or mouth frames where we kept the original facial frames when the iBug face landmark predictor failed to make accurate mouth or lip detection.

\section{EEG Feature Dimension Reduction Algorithm Details}
\label{sec:print}

The 155 EEG feature space was reduced to a dimension of 30 using non-linear principal component analysis. We used kernel PCA \cite{mika1999kernel} with polynomial kernel of degree 3 to perform non-linear PCA. Cumulative explained variance plots were used to get an idea to estimate the optimal dimension  \cite{krishna20}.
The non-linear dimension reduction was performed to further denoise the EEG feature space.

\section{Results}
\label{sec:page}
We used root mean square error (RMSE) computed between the predictions during test time and ground truth from test set as the performance metric to evaluate the model for each of the seven subjects. The obtained results are described in Figure 4. 
For predicting video from EEG features, subject 4 demonstrated lowest RMSE value of 12.3 during test time whereas for predicting EEG features from video, subject 1 demonstrated lowest RMSE value of 108.3 during test time.

The Figure 5 shows a video face frame for subject 1 from test set and Figure 6 shows the corresponding predicted facial video frame from the input EEG features during test time. It is clear from Figure 6 that only very broad characteristics of the facial video frame were observed during prediction during test time. The boundary of the face can be observed in Figure 6. This might be the first step towards the final goal of predicting high quality video frames from EEG as our model was able to learn very broad characteristics of facial video frames from input EEG features. 
\begin{figure}[h]
\begin{center}
\includegraphics[height=6.5cm,width=0.3\textwidth,trim={0.1cm 0.1cm 0.1cm 0.1cm},clip]{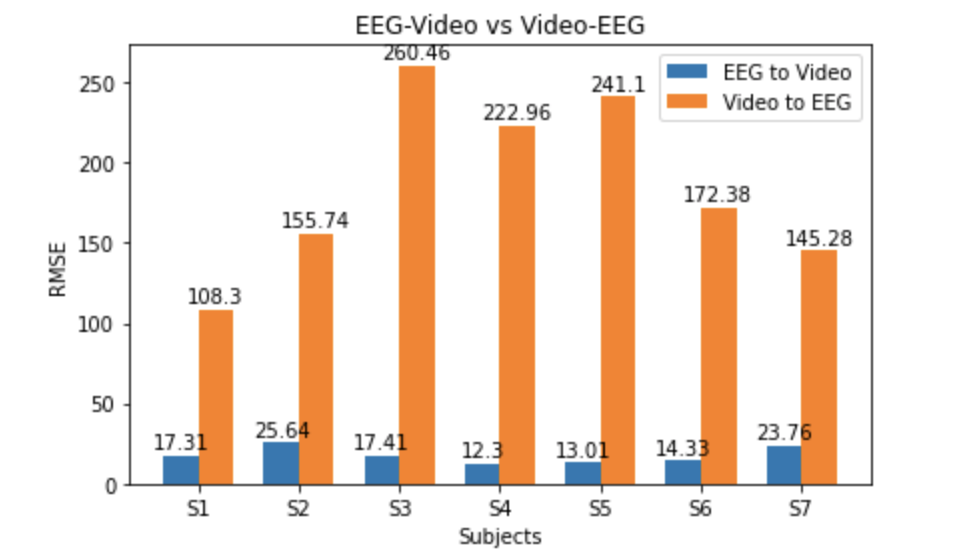}
\caption{Test time results} 
\label{1vsall}
\end{center}
\end{figure}

\begin{figure}[h]
\begin{center}
\includegraphics[height=5.5cm,width=0.3\textwidth,trim={0.1cm 0.1cm 0.1cm 0.1cm},clip]{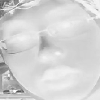}
\caption{ground truth video face frame from subject 1 test set} 
\label{1vsall}
\end{center}
\end{figure}

\begin{figure}[h]
\begin{center}
\includegraphics[height=5.5cm,width=0.3\textwidth,trim={0.1cm 0.1cm 0.1cm 0.1cm},clip]{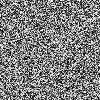}
\caption{corresponding predicted facial frame for subject 1 during test time} 
\label{1vsall}
\end{center}
\end{figure}

\section{Conclusions and Future work}
\label{sec:illust}
In this paper we explored predicting gray scale facial or lip video frames from input EEG features as well as predicting EEG features from input gray scale video frames using deep learning models. During test time we observed that our model was able to learn very broad characteristics of the facial video frames from input EEG features. Our results might be the first step towards the final goal of synthesizing high quality video from input EEG features. 

For future work we would like to improve our current results by training the models with a larger data set. It might also be worth exploring the use of generative adversarial networks (GAN) \cite{goodfellow2014generative} to solve these problems provided a larger data set is available to train the GAN model. In this work we observed poor performance when we performed experiments using GAN.

\section{Acknowledgements}
\label{sec:foot}
We would like to thank Kerry Loader and Rezwanul Kabir from Dell, Austin, TX for donating us the GPU to train the CTC model used in this work.




\bibliographystyle{IEEEbib}
\bibliography{strings,refs}

\end{document}